\documentclass{article}
\usepackage{spconf,amsmath,graphicx}
\usepackage{graphicx}
\usepackage{booktabs}
\usepackage{multirow}
\usepackage{algpseudocode}
\usepackage{amsmath}
\usepackage{physics}
\usepackage{setspace}
\usepackage{xcolor}
\usepackage{amssymb}

\newcommand{\matr}[1]{#1} 
\newcommand{\vect}[1]{\vb*{#1}}

\title{Exploring Attention Mechanisms for Multimodal Emotion Recognition in an Emergency Call Center Corpus}

\name{Théo Deschamps-Berger$^{\star}$ \qquad Lori Lamel$^{\star}$ \qquad Laurence Devillers$^{\dagger}$}
\address{$^{\star}$ LISN - CNRS, Paris-Saclay University, Orsay, France\\
$^{\dagger}$ LISN - CNRS, Sorbonne University, Orsay, France}
\begin{document}
%
\maketitle
\begin{abstract}

The emotion detection technology to enhance human decision-making is an important research issue for real-world applications, but real-life emotion datasets are relatively rare and small. The experiments conducted in this paper use the CEMO, which was collected in a French emergency call center. Two pre-trained models based on speech and text were fine-tuned for speech emotion recognition. Using pre-trained Transformer encoders mitigates our data's limited and sparse nature. This paper explores the different fusion strategies of these modality-specific models. In particular, fusions with and without cross-attention mechanisms were tested to gather the most relevant information from both the speech and text encoders. We show that multimodal fusion brings an absolute gain of 4-9\% with respect to either single modality and that the Symmetric multi-headed cross-attention mechanism performed better than late classical fusion approaches. Our experiments also suggest that for the real-life CEMO corpus, the audio component encodes more emotive information than the textual one.
%





%

\end{abstract}


\vspace*{-.100cm}
\section{Introduction}

In an emergency call center, an agent has to make quick decisions that can have great consequences. Automatic emotion recognition is a key component of human decision makings. The corpus used in this research (CEMO) was collected in a French emergency call center. In application-specific data, emotions are scarce accounting from 10\% (banking context) \cite{hardyMultilayerDialogueAnnotation2003} to 30\% (emergency call context) of speech turns/segments \cite{devillersChallengesReallifeEmotion2005a}. In this study we used and fine-tuned two pre-trained Transformer encoders, trained on French data: LeBenchmark wav2vec2 \cite{evainLeBenchmarkReproducibleFramework2021a} which is based on the wav2vec2 model designed to be trained on audio \cite{baevskiWav2vecFrameworkSelfSupervised2020b}; and FlauBERT \cite{leFlauBERTUnsupervisedLanguage2020a}, trained on text and based on the BERT model \cite{devlinBERTPretrainingDeep2019b}. These models have demonstrated their capabilities to produce relevant representations for several downstream tasks \cite{wagnerDawnTransformerEra2022, acheampongTransformerModelsTextbased2021}. 
Transformers encoder-decoder \cite{vaswaniAttentionAllYou2017a} have triggered novel approaches, thanks to the self-attention mechanism \cite{chengLongShortTermMemoryNetworks2016a} and to the cross-attention mechanism \cite{bahdanauNeuralMachineTranslation2015} that mixes two different embedding sequences which may come from different modalities. The attention mechanism has been used in speech emotion recognition with frame-level speech features \cite{xieSpeechEmotionClassification2019} and with knowledge transfer \cite{zhaoSelfattentionTransferNetworks2021}. The non-sequential characteristic of the attention mechanism enables Transformer to better capture long dependencies compared to Long-Short Term Memory. Specifically, Transformer encoders have been widely used to create domain-relevant representations in sequence processing. They were naturally first used on Written Language Processing tasks, particularly with language representation models such as BERT \cite{devlinBERTPretrainingDeep2019b} designed to serve as a pre-trained core model. It is trained in a self-supervised mode on huge amounts of data with the aim of being able to fine-tune it to specific applications and low-resource domains. In our case, the data are telephone conversations (speech and transcriptions) between call center agents and patients or relatives. In this real data recording context, data exhibiting emotion are rare and often have a wide diversity in the manner it is expressed: sometimes in the voice characteristics (prosody, fluency), in the choice of words, or a combination of both.
Multimodal deep learning is well-suited to handling this problem and has garnered much interest for emotion detection \cite{baltrusaitisMultimodalMachineLearning2019,tzirakisSpeechEmotionRecognition2021,tarantinoSelfAttentionSpeechEmotion2019, hanEmoBedStrengtheningMonomodal2021}: The paralinguistic representation provides low-level emotion cues, and the semantic features bring content and context to the sentence. The flexibility of deep learning systems supports several multi-modal fusion levels: early \cite{wimmerLowLevelFusionAudio2008, bussoAnalysisEmotionRecognition2004}, intermediate \cite{hanStrengthModellingRealWorld2017, delbrouckModulatedFusionUsing2020c, liuEfficientLowrankMultimodal2018}, and late \cite{zengSurveyAffectRecognition2009, hoMultimodalApproachSpeech2020a, tzirakisEndtoendMultimodalAffect2021}. 
In this paper, we focus on late fusion, which introduces the multimodal information in the later layers of the network, allowing the earlier layers to specialize the uni-modal pattern learning and extraction. We compare three late fusions strategies: the first is a score average of each specific-modality encoder (Score fusion)  \cite{zengSurveyAffectRecognition2009}; the second is a shared neural network fed by the concatenation of both encoders outputs (Concatenation fusion) \cite{hanStrengthModellingRealWorld2017,hanPredictionbasedLearningContinuous2017} and the third is a Symmetric multi-head cross-attention fusion which takes in context the semantic and the paralinguistic representations to weight the final output. A similar multi-head Symmetric cross-attention method was used in a multimodal transformer for acoustics and vision in \cite{nagraniAttentionBottlenecksMultimodal2021} and for speech emotion recognition with MFCC and GloVe embeddings \cite{sunMultimodalCrossSelfAttention2021}. 
This paper extends our previous studies on the adaptation of Transformer encoders for speech emotion recognition and on late multimodal fusions \cite{deschamps-bergerEndtoEndSpeechEmotion2021, deschamps-bergerInvestigatingTransformerEncoders2022}. To the best of our knowledge, the use of Symmetric cross-attention powered by fine-tuned pre-trained encoders for speech emotion recognition has not been realized before. The rest of the paper is organized as follows. 
The next section explains how the encoders are fine-tuned for emotion recognition.
Then, we detail the fusion strategies, and present the CEMO corpus and the experimental results.

\vspace*{-.200cm}
\section{Fine-tuned pre-trained encoders}

\begin{figure}
  \centering
  \caption{ 
  Parameter fine-tuning of the wav2vec2 and FlauBERT transformer layers. The Convolutional and Embedding layers are frozen.
  The wav2vec2 and the FlauBERT produce paralinguistic ($H_p$) and semantic representations ($H_s$), respectively.}
  \includegraphics[width=1\linewidth]{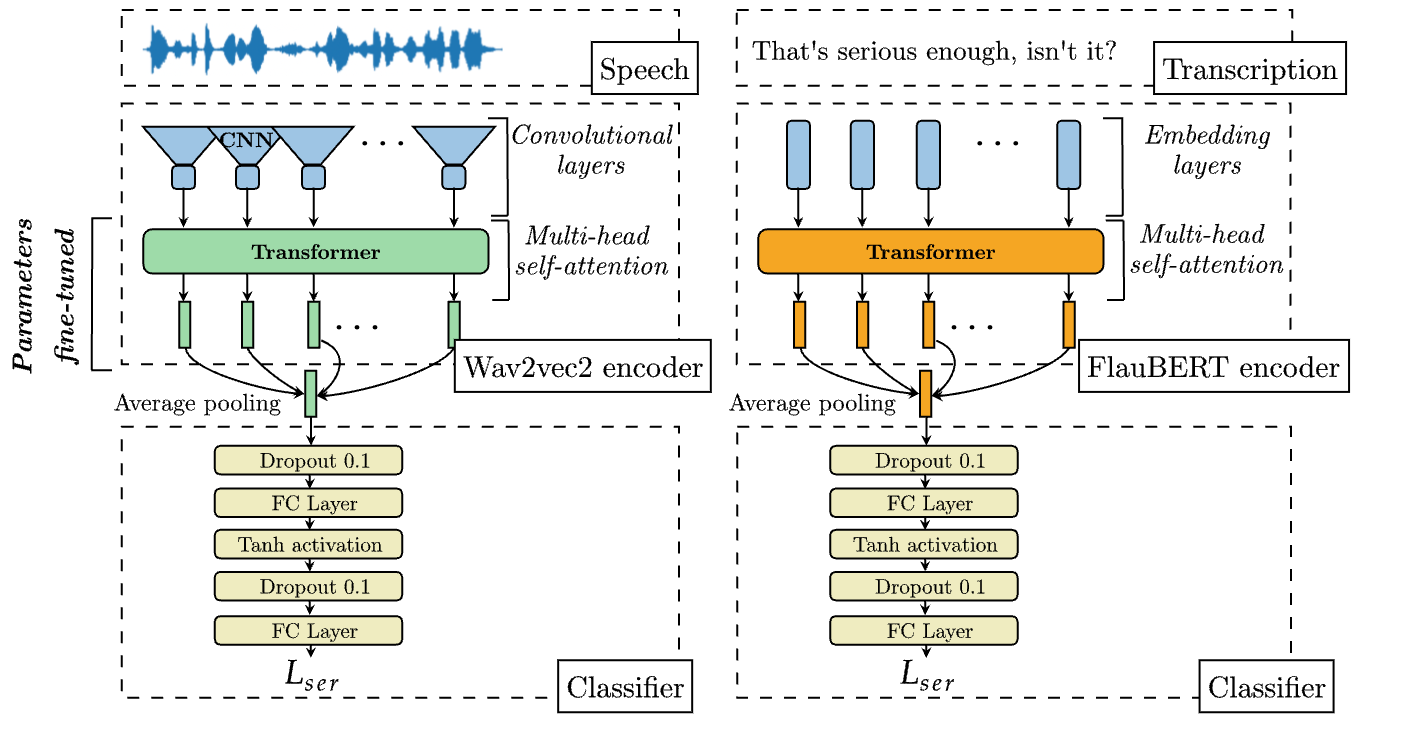}
  \vspace*{-.75cm}
  \label{fig:fine-tuning}
\end{figure}

\vspace*{-1ex}
In prior results \cite{deschamps-bergerInvestigatingTransformerEncoders2022},  several pre-trained models for speech emotion recognition were adapted with the CEMO corpus \cite{devillersChallengesReallifeEmotion2005a}. 
We selected among the available models the leBenchmark model (Wav2Vec2-FR-3K) \cite{evainLeBenchmarkReproducibleFramework2021a} trained on 3k hours of French and the FlauBERT model \cite{leFlauBERTUnsupervisedLanguage2020a} trained on 24 French subcorpora (71Gb of text). This selection is justified by the performances on our CEMO corpus \cite{deschamps-bergerInvestigatingTransformerEncoders2022}. The wav2vec2-FR-3K model includes in its training database some spontaneous dialogues, telephone-recorded data, and emotional content, which have similar characteristics to the CEMO corpus. The FlauBERT model applies a basic tokenizer for French to the input sentence and embeds it with a vocabulary of 50K subword units built using the Byte Pair Encoding (BPE) algorithm \cite{sennrichNeuralMachineTranslation2016}. 
%
%
In order to adapt the wav2vec2-FR-3K and FlauBERT models, we added a classifier on top of them, adapted to our four emotional classes. The multi-head self-attention layers of the Transformer encoders were fine-tuned for speech emotion recognition with the CEMO corpus, as shown in Figure \ref{fig:fine-tuning}, making the assumption that the first layers (Convolutional and Embedding) are reliable for this task \cite{wagnerDawnTransformerEra2022, deschamps-bergerInvestigatingTransformerEncoders2022}.

\vspace*{-.200cm}
\section{Multimodal fusions strategies}

The multimodal fusion variants are applied to the outputs of both encoders, fine-tuned on speech or transcriptions, respectively. The wav2vec2 encoder produces a matrix composed of paralinguistic features per frame window $(\matr{H_p} \in \mathbb{R}^{d_{\text{frames}}\times d_{\text{model}}})$ while the FlauBERT encoder yields a matrix of semantic features per word $(\matr{H_s} \in \mathbb{R}^{d_{\text{subwords}}\times d_{\text{model}}})$. 
For the Base configuration the encoders have $d_{\text{model}}=768$ and $d_{\text{model}}=1024$ for the Large size.
We tested three methods to extract the most relevant information from both feature spaces.
\vspace*{-.145cm}
\subsection{Score fusion}

The output matrix of each encoder is averaged across the model dimension ($\text{mean}(\matr{H_p})$ and $\text{mean}(\matr{H_s})\in\mathbb{R}^{d_{\text{model}}}$). The Score fusion consists of two fully connected parallel layers processing the average outputs of the two encoders.
$$
\begin{aligned}
&\tilde{\vect{y}}_{p}=\sigma(tanh(\matr{W_p} \times \text{mean}(\matr{H}_p)+b_{p})), \\
&\tilde{\vect{y}}_{s}=\sigma(tanh( \matr{W_s} \times \text{mean}({\matr{H}}_{s})+b_{s})),
\end{aligned}
$$
where $\matr{W_p}$ and $\matr{W_s}\in \mathbb{R}^{n_{\text{classes}}\times d_{\text{model}}}$.
The softmax outputs generated by each classifier are then averaged by class.
\vspace*{-.145cm}
\subsection{Concatenation fusion}

The Concatenation fusion consists of shared fully connected layers fed by the concatenation of the encoder's output matrices and averaged across the model dimension.
$$
\begin{aligned}
&\vect{h_\text{concat}}=\left[\text{mean}(\matr{H}_p), \text{mean}(\matr{H}_s)\right]\\
&\tilde{\mathbf{y}}_{c}=\sigma(tanh(\matr{W}_{c} \times \vect{h_\text{concat}}+b_{c}))
\end{aligned}
$$
where $\matr{W}_{c} \in \mathbb{R}^{n_{\text{classes}}\times d_{\text{model}}}$.

\vspace*{-.145cm}
\subsection{Symmetric multi-head cross-attention fusion}

The attention algorithm relies on the idea that each element of an input matrix can be weighted by its importance in the context. To this end, the input matrix (which we call the Key matrix) is multiplied by a context matrix (denoted as Query) and then softmax and scaled to produce an attention matrix. This attention matrix is finally multiplied by the input matrix (called Value) to produce a matrix whose elements are weighted by the context; see equation \ref{equ:att}. In self-attention, the context matrix ($\matr Q$) is equal to the input matrix ($\matr K$, $\matr V$); in cross-attention, the context matrix ($\matr Q$)  comes from another modality. 
\begin{equation}
\label{equ:att}
\operatorname{Attention}(\matr Q,\matr K,\matr V)=\operatorname{softmax}\left(\frac{Q K^{T}}{\sqrt{d_{k}}}\right) V
\end{equation}
\indent where ${d_k}$ is the dimension of queries and keys.

\begin{figure}
    \centering
    \caption{
    {Multimodal system with Symmetric multi-head cross-attention fusion.}}
    \includegraphics[width=1\columnwidth]{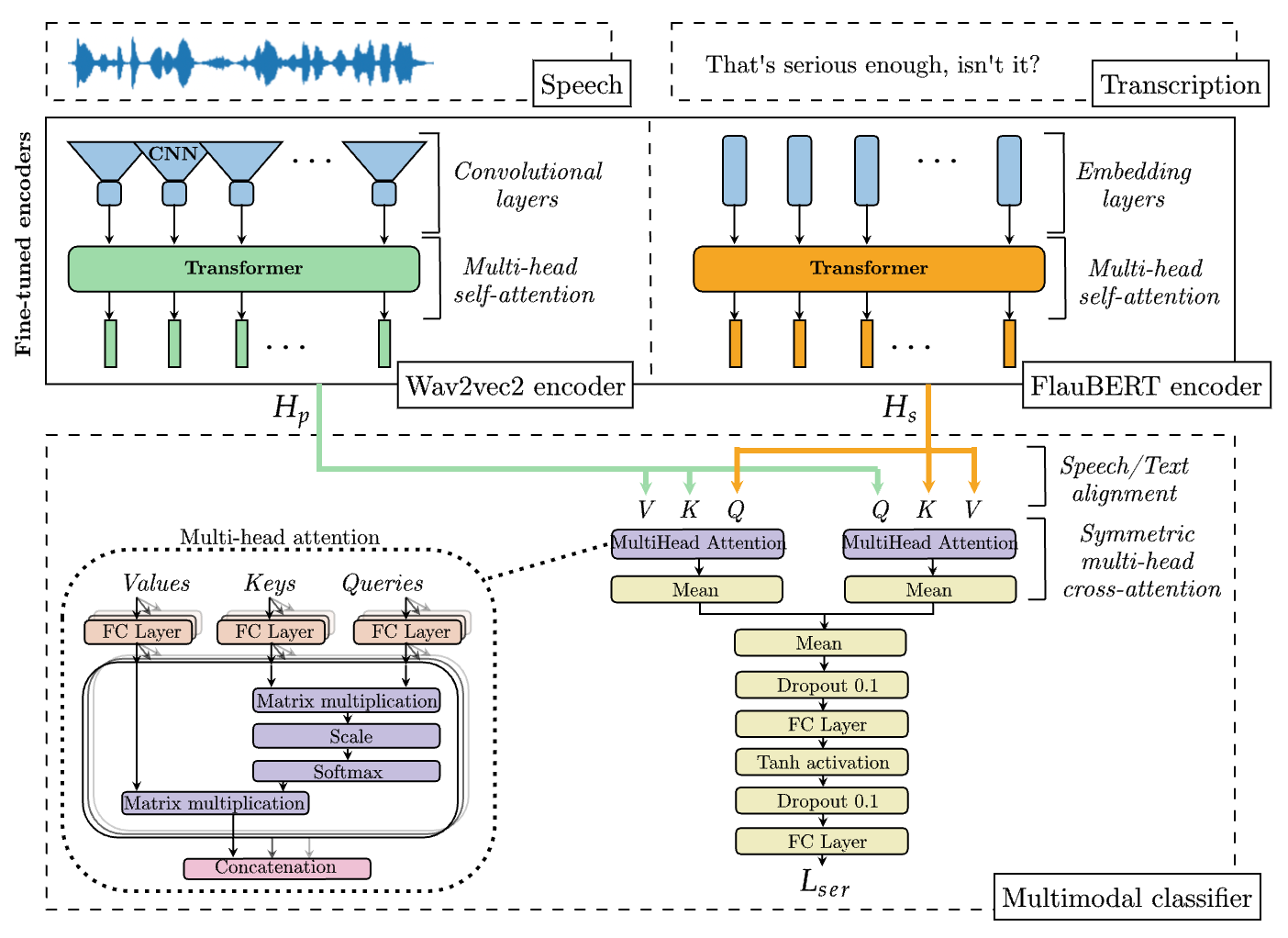}
    \vspace*{-.5cm}
    \label{fig:archi}
\end{figure}

\noindent Following Vaswani~ \cite{vaswaniAttentionAllYou2017a}, we used multi-head attention (eq.\, 2) in order to increase the number of attention mechanisms on different representation subspaces, as shown in Figure~\ref{fig:archi}. 
$$
\begin{aligned}
\operatorname{MHead}(Q, K, V) &=\operatorname{Concat}\left(\operatorname{head}_{1}, \ldots, \operatorname{head}_{\mathrm{h}}\right) W^{O} \hspace{0.52cm}\text{(2)}\\
\text { where head } &=\operatorname{Attention}\left(Q W_{i}^{Q}, K W_{i}^{K}, V W_{i}^{V}\right)
\end{aligned}
$$
\noindent And where the projections are parameter matrices: \\
$W_{i}^{Q} \in \mathbb{R}^{d_{\text {modcl }} \times d_{k}}, W_{i}^{K} \in \mathbb{R}^{d_{\text {model }} \times d_{k}}, W_{i}^{V} \in \mathbb{R}^{d_{\text {model }} \times d_{v}}$, $W^{O} \in \mathbb{R}^{h d_{v} \times d_{\text {model }} \text {. }} $ and $ d_v$ is the dimension of values.\\
In this work, we used 16 attention heads ($h=16$) for all fusions with Base model encoders and 32 heads ($h=32$) for Large encoders. 

To perform cross-attention, since the dimensions of the encoder outputs for wav2vec2 and FlauBERT are different, we experimented with two alignment methods. The first method is coarsely aligning the two embedding matrices to create equivalent matrices size. To this end, the number of frames in $H_p$ is divided by the number of subwords in $H_s$, and the aligned frames are averaged over each subword (entry \#subwords in Table \ref{tab:results}). In the second method, the number of frames associated with a subword depends on the number of characters in it, and in order to give more weight to longer subwords, the frames are summed rather than averaged. This is essentially equivalent to aligning at the character level and summing the frames in each character (entry \#characters in Table \ref{tab:results}).
\noindent We experimented with three fusions: Paralinguistic, Semantic, and Symmetric multi-head cross-attention fusions. The Paralinguistic and Semantic attention fusions consist of a multi-head attention with respectively: $(Q=H_s, K=H_p, V=H_p)$ and $(Q=H_p, K=H_s, V=H_s)$.

We propose a Symmetric multi-head cross-attention fusion, obtained by averaging the outputs of the two first multi-head cross-attention, as detailed in Figure~\ref{fig:archi}. We refer to these fusions as: Paralinguistic, Semantic, and Symmetric cross-attention in Table \ref{tab:results}.

\vspace*{-.200cm}
\section{Experimental results}

All experiments were assessed using a subset of the CEMO corpus described in the next subsection. 
The evaluation is performed on five-fold with a classical cross-speaker folding strategy that is speaker-independent between training, validation, and test sets. During each fold, system training is optimized on the best Unweighted Accuracy (UA) validation set. The outputs of each fold are combined for the final results.

\vspace*{-.145cm}
\subsection{Corpus}

\begin{table}
\centering
\caption{Details of the CEMO subset of speech signals and manual transcripts. ANG: Anger, FEA: Fear, NEU: Neutral, POS: Positive, Total: Total number of segments.}
\label{tab:CEMOdetails}
\resizebox{\columnwidth}{!}{
\begin{tabular}{lccccc} 
\toprule
\textbf{CEMO$_s$}           & \textbf{ANG}  & \textbf{FEA}  & \textbf{NEU}  & \textbf{POS}  & \textbf{Total}  \\ 
\midrule
\#Speech seg.    & 1056 & 1056 & 1056 & 1056 & 4224   \\
\#Speakers           & 149  & 537  & 450  & 544  & 812    \\
\#Dialogues          & 280  & 504  & 425  & 516  & 735    \\
Duration (min, s)    & 39   & 52   & 49   & 20   & 160    \\
Duration (mean, s)  & 2.2  & 2.9  & 2.8  & 1.1  & 2.3    \\
Vocabulary size    & 1146 & 1500 & 1150 & 505  & 2600   \\
Avg.\ word count & 9.3  & 11.9 & 7.9  & 3.8  & 8.2    \\
\bottomrule
\end{tabular}
}
\end{table}

The speech corpus was collected in a French emergency call center \cite{devillersChallengesReallifeEmotion2005a}. Data preparation is crucial to achieving good performance and robustness; here, we describe the selection of a balanced subset of the CEMO data used for model training, validation and test, as detailed in Table \ref{tab:CEMOdetails}. The selected CEMO subset (2h40) is comprised of 4224 segments, with durations ranging from 0.4 to 7.5 seconds equally distributed over the four main emotion classes: {\sc anger, fear, neutral} and {\sc positive}. To this end, the Fear and Neutral classes were subsampled, maintaining 1056 samples for each class for which the annotators agreed while prioritizing a broad representation of speaker diversity. The minority class Anger was completed with segments from the agents. Since the patient Anger class is composed of over 85\% fine-grained segments of annoyance and impatience, the selected segments in the agents were selected from the same fine-grained labels. As seen in Table \ref{tab:CEMOdetails}, there are fewer speakers for Anger compared to the other classes. It can be noted in the Positive class has the most significant number of speakers and dialogues, potentially being richer and more heterogeneous than the other classes. Manual transcriptions of the CEMO corpus were performed by two coders, with guidelines similar to those proposed in the Amities project \cite{hardyMultilayerDialogueAnnotation2003}. 
The transcriptions contain about 2499 nonspeech markers, primarily pauses, breath, and other mouth noises. The vocabulary size is 2.6k, with a mean and median of about 10 words per segment (min 1, max 47).

\vspace*{-.345cm}
\subsection{Results}

\begin{table}
\centering
\caption{
{Unimodal results with fine-tuning for speech emotion recognition with the CEMO corpus. Configuration (Config.), Fine-tuned parameters (Fine-tuned\,p.), UA: \%}}
\label{tab:ft}
\resizebox{\linewidth}{!}{%
\begin{tabular}{llllllll} 
\toprule
\textbf{Models}                  & \textbf{Config.} & \multicolumn{1}{c}{\textbf{Fine-tuned\,p.}} & \textbf{ANG} & \textbf{FEA} & \textbf{NEU} & \textbf{POS} & \multicolumn{1}{c}{\textbf{Total}}  \\ 
\midrule
\multirow{2}{*}{Wav2Vec2-FR-3K~\cite{evainLeBenchmarkReproducibleFramework2021a}} & Base             & 90 M                                       & 70.4         & 70.5         & 74.8         & 83.2         & 74.7                                \\
                                 & Large            & 311 M                                      & 71.6         & 69.8         & 71.7         & 88.4         & 75.4                                \\ 
\midrule
\multirow{2}{*}{FlauBERT~\cite{leFlauBERTUnsupervisedLanguage2020a}}        & Base             & 85 M                                       & 57.9         & 64.9         & 74.1         & 83.5         & 70.1                                \\
                                 & Large            & 303 M                                      & 61.0         & 60.7         & 73.2         & 85.6         & 70.1                                \\
\bottomrule
\end{tabular}
}
\end{table}

\begin{table}
\centering
\caption{Multimodal fusions comparative experiments. Configuration (Config.), Trainable parameters (Train p.), UA: \%}
\label{tab:results_score_cat}
\resizebox{\linewidth}{!}{%
\begin{tabular}{llcccccc} 
\toprule
\textbf{Fusion}                & \textbf{Config.} & \textbf{Train.\,p.} & \textbf{ANG}  & \textbf{FEA}  & \textbf{NEU}  & \textbf{POS}  & \textbf{Total}  \\ 
\midrule
\multirow{2}{*}{Score}         & Base             & 2 M                & 69.9          & 73.8          & 75.7          & 84.9          & 76.1            \\
                               & Large   & 4 M       & 79.7 & 70.4 & 77.0 & 89.0 & 79.0   \\ 
\midrule
\multirow{2}{*}{Concatenation} & Base    & 2 M       & 70.5 & 73.8 & 75.7 & 85.0 & 76.3   \\
                               & Large            & 4 M                & 78.7          & 69.4          & 77.2          & 88.3          & 78.4            \\
\bottomrule
\end{tabular}
    \vspace*{-.5cm}
}
\end{table}

Table~\ref{tab:ft} provides the unimodal results, which serve as a baseline for this study. During training, we use cross-entropy loss, Adam optimization \cite{kingmaAdamMethodStochastic2015}, with a fixed learning rate of $10^{-5}$, and a mini-batch of size 8. For encoder fine-tuning, due to a large number of parameters, we used gradient norm clipping of $1$, to facilitate learning stability, improving our prior results for Large encoders \cite{deschamps-bergerInvestigatingTransformerEncoders2022}. 

\begin{table}
\centering
\caption{Multimodal multi-head cross-attention fusion comparison. Configuration (Config.), Trainable parameters (Train p.), UA: \%}
\label{tab:results}
\resizebox{\linewidth}{!}{%
\begin{tabular}{ccccccccc} 
\toprule
\textbf{Fusion}                                                                    & \textbf{Alignment}                     & \textbf{Config.} & \textbf{Train.\,p.} & \textbf{ANG}  & \textbf{FEA}  & \textbf{NEU}  & \textbf{POS}  & \textbf{Total}  \\ 
\midrule
\multirow{4}{*}{\begin{tabular}[c]{@{}c@{}}Paralinguistic\\cross-attention\end{tabular}} & \multirow{2}{*}{\#subwords}   & Base             & 2 M                & 69.6          & 74.5          & 74.1          & 84.6          & 75.7            \\
                                                                                   &                               & Large            & 4 M                & 80.2          & 71.2          & 74.2          & 87.3          & 78.2            \\ 
\cmidrule{2-9}
                                                                                   & \multirow{2}{*}{\#characters} & Base             & 2 M                & 69.5          & 74.1          & 74.1          & 84.1          & 75.5            \\
                                                                                   &                               & Large            & 4 M                & 79.1          & 73.4          & 74.2          & 87.1          & 78.5            \\ 
\midrule
\multirow{4}{*}{\begin{tabular}[c]{@{}c@{}}Semantic \\cross-attention\end{tabular}}      & \multirow{2}{*}{\#subwords}   & Base             & 2 M                & 69.8          & 71.0          & 74.1          & 86.8          & 75.5            \\
                                                                                   &                               & Large            & 4 M                & 78.3          & 71.1          & 75.9          & 87.9          & 78.3            \\ 
\cmidrule{2-9}
                                                                                   & \multirow{2}{*}{\#characters} & Base             & 2 M                & 69.2          & 69.7          & 74.3          & 85.8          & 74.8            \\
                                                                                   &                               & Large            & 4 M                & 77.7          & 69.7          & 73.8          & 86.8          & 77.0            \\ 
\midrule
\multirow{4}{*}{\begin{tabular}[c]{@{}c@{}}\textbf{Symmetric} \\\textbf{cross-attention}\end{tabular}}     & \multirow{2}{*}{\#subwords}   & Base    & 5 M       & 70.2 & 75.2 & 75.5 & 84.8 & 76.4   \\
                                                                                   &                               & Large            & 8 M                & 81.9          & 70.5          & 77.0          & 87.8          & 79.3            \\ 
\cmidrule{2-9}
                                                                                   & \multirow{2}{*}{\textbf{\#characters}} & Base             & 5 M                & 68.6          & 74.5          & 75.9          & 84.8          & 75.9            \\
                                                                                   &                               & \textbf{Large}   & \textbf{8 M}       & \textbf{82.4} & \textbf{71.5} & \textbf{75.6} & \textbf{87.9} & \textbf{79.4}   \\
\bottomrule
\end{tabular}
}
\end{table}

Both sized audio encoder configurations give an absolute gain of 4-5\% with respect to the fine-tuned FlauBERT encoders, (c.f. Table \ref{tab:ft}). This result suggests that, for our real-life CEMO corpus, the paralinguistic model encodes more emotive information than the textual one. The baseline fusion methods, Multimodal Score and Concatenation, with both encoder sizes (Base and Large) outperform the single modality encoders (compare Tables \ref{tab:ft} and \ref{tab:results_score_cat}). 
In Table \ref{tab:results}, we explored the cross-attention mechanisms with three attention-based fusion strategies for the last fusion. The Paralinguistic cross-attention fusion almost always performed better than the Semantic fusion with both alignment strategies. The better results of the unimodal wav2vec2 compared to the FlauBERT encoder show that the paralinguistic encoder produces relevant features for speech emotion recognition. In addition to including acoustic information, the wav2vec2 encoder also implicitly includes lexical information. The encoder size influences the multimodal fusion performance. The Base and Large unimodal fine-tuned encoders have similar results for the Paralinguistic encoders (around 0.7\% difference in Table \ref{tab:ft}), and the same results with the Semantic encoders. However, the Large configuration yielded a 2.9\% improvement over the Base model for the Symmetric cross-attention fusion method with alignment based on subwords, as shown in Table \ref{tab:results}. For a given condition (fusion method and model size), the \#subwords alignment method generally performs slightly better than the \#characters alignment. 

The CEMO corpus contains data samples from 812 speakers, expressive voices, and nonspeech markers comprising sentences' most frequent emotional cues. The corpus has a vocabulary of only 2600 different words, of which only a few predict emotions. In particular, Fear and Anger are poorly detected by the Semantic models.
These patterns make the fusion more complex. The Symmetric cross-attention fusion outperformed single modality encoders with both alignment and encoder sizes and performed slightly better than the classical fusion Score and Concatenation fusions. The best performance of the Symmetric cross-attention multimodal fusion is 79.3\% UA compared with 74.5\% UA (unimodal fine-tuned wav2vec2) or 70.1\% UA (unimodal fine-tuned FlauBERT).

\vspace*{.300cm}
\section{Conclusion}

This paper explored different fusion mechanisms in a multimodal context with real-life recordings from a French emergency call center. In this application, the agent aims to gather as much objective information about the situation as possible to make an informed decision. At the same time, the patient is more concerned with explaining the emergency,  their pain, or their anxiety. These characteristics make the detection of emotion from speech very complex, as the emotion cues can be present in audio and/or lexical information.
The multimodal fusions were powered by unimodal pre-trained encoders that were fine-tuned for speech emotion recognition using the CEMO corpus. Three primary multimodal fusions were tested; Score, Concatenation, and Symmetric cross-attention. The best emotion recognition results for CEMO were obtained using a Symmetric fusion associated with an adequate alignment. Further experiments 
will be carried out
using transcriptions produced by an automatic speech recognizer instead of manual references. 

\vspace*{-.145cm}
\subsection*{Ethics and reproducibility}
The use of the CEMO database or any subsets of it, carefully respected ethical conventions and agreements ensuring the anonymity of the callers. All the experiments were carried out using Pytorch on GPUs (Tesla V100 with 32 Gbytes of RAM). To ensure the reproducibility of the runs, we set a random seed to 0 and prevent our system from using non-deterministic algorithms. This work was performed using HPC resources from GENCI–IDRIS (Grant 2022-AD011011844R1).

\pagebreak

\bibliographystyle{IEEE}
\setstretch{0.80}
\bibliography{icassp23}

\end{document}